\pdfoutput=1

\documentclass[11pt]{article}

\usepackage[]{acl2021}

\usepackage{times}
\usepackage{latexsym}

\usepackage[T1]{fontenc}

\usepackage[utf8]{inputenc}

\usepackage{microtype}

\usepackage{makecell}

\usepackage{microtype}

\usepackage{graphicx}
\usepackage{paralist}
\usepackage{url}
\usepackage{xspace}
\usepackage{amsfonts,amsmath}
\usepackage{booktabs}
\usepackage{setspace}
\usepackage{multirow}
\usepackage{multicol}
\usepackage{subfigure}
\usepackage{hyperref}
\usepackage{xcolor}
\usepackage{amssymb}
\usepackage{pifont}
\usepackage{color, colortbl}
\usepackage{bbm}
\usepackage{wrapfig}
\usepackage{lipsum}
\usepackage{enumitem}
\usepackage{paralist, tabularx}
\usepackage{balance}

\newcommand*\colourTrue[1]{%
  \expandafter\newcommand\csname #1True\endcsname{\textcolor{#1}{\ding{52}}}%
}
\colourTrue{teal}
\colourTrue{red}

\newcommand*\colourFalse[1]{%
  \expandafter\newcommand\csname #1False\endcsname{\textcolor{#1}{\ding{56}}}%
}
\colourFalse{green}
\colourFalse{red}

\title{\textls[-10]{Retrieval Augmentation for Commonsense Reasoning: A Unified Approach}}

\author{Wenhao Yu$^{1}$, Chenguang Zhu$^{2}$, Zhihan Zhang$^{1}$, Shuohang Wang$^{2}$, \\ \bf Zhuosheng Zhang$^{3}$, Yuwei Fang$^{2}$, Meng Jiang$^{1}$\\
$^1$University of Notre Dame, Indiana, USA \\ $^2$Microsoft Cognitive Services Research, Washington, USA \\ $^3$Shanghai Jiaotong University, Shanghai, China  \\
{$^1$\tt \{wyu1, zzhang23, mjiang2\}@nd.edu};  \\
{$^2$\tt \{chezhu, shuow, yuwfan\}@microsoft.com}; 
{$^3$\tt zhangzs@sjtu.edu.cn}
}

\begin{document}
\maketitle

\begin{abstract}
A common thread of retrieval-augmented methods in the existing literature focuses on retrieving encyclopedic knowledge, such as Wikipedia, which facilitates well-defined entity and relation spaces that can be modeled.
However, applying such methods to commonsense reasoning tasks faces two unique challenges, i.e., the lack of a general large-scale corpus for retrieval and a corresponding effective commonsense retriever.
In this paper, we systematically investigate how to leverage commonsense knowledge retrieval to improve commonsense reasoning tasks.
We proposed a unified framework of \underline{\textbf{R}}etrieval-\underline{\textbf{A}}ugmented \underline{\textbf{Co}}mmonsense reasoning (called \textsc{RACo}), including a newly constructed commonsense corpus with over 20 million documents and novel strategies for training a commonsense retriever. We conducted experiments on four different commonsense reasoning tasks. Extensive evaluation results showed that our proposed \textsc{RACo} can significantly outperform other knowledge-enhanced method counterparts, achieving new SoTA performance on the CommonGen\footnote{\url{https://inklab.usc.edu/CommonGen/leaderboard.html}} and CREAK\footnote{\url{https://www.cs.utexas.edu/~yasumasa/creak/leaderboard.html}} leaderboards. Our code is available at \url{https://github.com/wyu97/RACo}.

\end{abstract}

\section{Introduction}
\label{sec:introduction}
Recent work has shown that scaling language models with considerably more data and parameters, such as 
GPT3-175B~\cite{brown2020language}, could drive significant advances in commonsense reasoning tasks. 
Nevertheless, such models make predictions by only ``looking up information” stored in their parameters, making it difficult to determine what knowledge is stored or has been already forgotten by the neural network~\cite{guu2020realm}.
Besides, storage space is limited by the size of the neural network. In order to memorize more world knowledge, one must train ever-larger networks, which can be prohibitively expensive and slow.

\begin{table*}[t]
\begin{center}
\setlength{\tabcolsep}{1mm}{\scalebox{0.81}{\begin{tabular}{l|ccccc}
\toprule
& \textsc{\textbf{RACo}} & \textsc{AristoRoBERTa} & \textsc{Re-T5} & \textsc{KFCNet} & \textsc{OpenCSR} \\
& \textbf{(this work)} & \cite{mihaylov2018can} & \cite{wang2021retrieval} & \cite{li2021kfcnet} & \cite{lin2021differentiable} \\
\midrule
Number of corpus types & \textbf{3} & 1 & 1 & 1 & 1 \\
Number of commonsense tasks & \textbf{4} & 1 & 1 & 1 & 1  \\
Number of docs for retrieval & \textbf{20M} & 5K & 0.8M & 0.8M & 1M \\
\bottomrule
\end{tabular}}}
\end{center}
\vspace{-0.2in}
\caption{Comparison of \textsc{RACo} to a few recent commonsense retrieval works in the field. Our work provides a more comprehensive and larger-scale multi-source commonsense corpus that can generalize to various tasks.}
\label{tab:comparison}
\end{table*}

The solution that may seem obvious at first glance is to grant language models free access to open-world sources of commonsense knowledge in a plug-and-play manner, instead of memorizing all world knowledge.
To achieve this capability, language models must be able to \textit{retrieve} relevant commonsense knowledge from an unbounded set of situations.
Then, the language models can leverage the input text, as well as the retrieved information to produce the desired output.

Compared with the large-scale language model counterparts, e.g., \textsc{Unicorn}~\cite{lourie2021unicorn}, retrieval-augmented methods have three remarkable advantages: first, the knowledge is not stored implicitly in the model parameters, but is explicitly acquired in a plug-and-play manner, leading to great scalability; second, the paradigm generates text based on some retrieved references, which
alleviates the difficulty of generating from scratch~\cite{li2022survey}; third, knowledge corpus can be constantly edited and updated by experts, making the model aware of the latest information. 
Besides, compared with knowledge graph inference model counterparts, e.g., QA-GNN~\cite{yasunaga2021qa}, retrieval-augmented methods allow more flexibility in accessing and using knowledge from different sources, because of the nature of commonsense knowledge, which cannot all be contained in a single knowledge graph defined by a certain schema~\cite{yu2020survey}.

A common thread of retrieval-augmented methods in the existing literature focuses on retrieving encyclopedic knowledge such as Wikipedia, which lends itself to a well-defined space of entities and relations that can be modeled~\cite{karpukhin2020dense,lewis2020retrieval,yu2022kg}.
However, retrieval-augmented methods for commonsense reasoning have been rarely studied in the literature.
In this paper, we propose a unified framework of \underline{\textbf{R}}etrieval-\underline{\textbf{A}}ugmented \underline{\textbf{Co}}mmonsense reasoning (\textsc{RACo}) to solve various commonsense tasks.
\textsc{RACo} first retrieves relevant commonsense documents from a large-scale corpus, then combines the input text with the retrieved documents to produce the desired output.
However, there are two main challenges in training a \textsc{RACo} model.

The first challenge to address is \textit{what} commonsense knowledge to retrieve.
Different from encyclopedic knowledge used in open-domain QA tasks, commonsense knowledge is very diverse, containing everyday events and their effects,
facts about beliefs and desires,
and properties of objects in human’s daily life.
Since commonsense involves various aspects including human interaction and object properties in everyday life, we collected a over 20 million commonsense documents collection from both open-domain knowledge sources (e.g., OMCS) that cover multiple domains of commonsense, and domain-specific sources (e.g., ATOMIC) that focus on particular commonsense types.

The second challenge is to address \textit{how} to retrieve relevant commonsense knowledge from the corpus. Different from training a dense retriever on Wikipedia~\cite{karpukhin2020dense}, the heuristic of taking ``documents containing correct answers'' as positive candidates cannot be used because the output answer in commonsense reasoning tasks is usually not a substring of retrieved documents. For example, in binary question answering, the answer is \textit{True} or \textit{False} but it does not appear in the retrieved documents. 
Therefore, we propose novel strategies to construct question-document pairs for commonsense dense retriever training. 

Overall, our main contributions in this work can be summarized as follows:
\begin{enumerate}[wide=5pt, itemsep=-0.5ex, topsep=-2pt,]
    \item We collected and publicized a collection of over 20 million documents from three knowledge sources for commonsense knowledge retrieval.
    \item We presented a unified framework of \underline{\textbf{R}}etrieval- \underline{\textbf{A}}ugmented \underline{\textbf{Co}}mmonsense reasoning (\textsc{RACo}), and proposed novel strategies for training a strong commonsense knowledge retriever. 
    \item We evaluated our \textsc{RACo} on four types of commonsense reasoning tasks. Our experiments showed \textsc{RACo} can significantly outperform other knowledge-enhanced counterparts, achieving new SoTA on CommonGen and CREAK leaderboards.
\end{enumerate}

\section{Related Work}
\label{sec:related}
Though large-scale language models yield state-of-the-art performance on many commonsense reasoning tasks, their pre-training objectives do not explicitly guide the models to reason with commonsense knowledge such as the relation and composition of daily concepts in our lives~\cite{zhou2021pre}, leading to unsatisfactory performance in many real-world scenarios~\cite{talmor2021commonsenseqa,zhu2022knowledge}. 
Existing work has mainly explored two directions to improve their commonsense reasoning ability. 
The first is to pre-train or post-train a language model on commonsense corpora~\cite{bosselut2019comet,lourie2021unicorn,zhou2021pre}. 
When the commonsense materials are appropriately selected, this simple strategy could demonstrate significantly superior performance than vanilla pre-trained language models~\cite{zhou2021pre}.
Notable methods include COMET~\cite{bosselut2019comet}, CALM~\cite{zhou2021pre}, \textsc{Unicorn}~\cite{lourie2021unicorn}, etc. 
Nonetheless, these methods still suffer from the same drawbacks as the pre-trained language models introduced in $\S$\ref{sec:introduction}.
The second is to explicitly introduce external knowledge from commonsense knowledge graphs to augment the limited textual information. ~\cite{lin2019kagnet,ji2020language}.
A KG often provides comprehensive and rich entity features and relations so models can easily traverse links to discover how entities are interconnected to express certain commonsense knowledge.
Notable methods include KagNet~\cite{lin2019kagnet}, GRF~\cite{ji2020language}, QA-GNN~\cite{yasunaga2021qa}, GreaseLM~\cite{zhang2022greaselm}, etc.
However, commonsense knowledge lies at an unbounded set of facts and situations that usually cannot be covered by a single knowledge graph defined by a certain schema. Reasoning over multiple knowledge graphs is a challenging task.

Retrieval-augmented method is a new learning paradigm that fuses pre-trained language models and traditional information retrieval techniques~\cite{lewis2020retrieval}.
A few recent methods have explored retrieving \textit{in-domain} commonsense documents from a task-relevant corpus to improve commonsense reasoning performance~\cite{mihaylov2018can,wang2021retrieval,li2021kfcnet}.
We provide a detailed comparison in Table \ref{tab:comparison}. 
Different from existing methods that focus on retrieving knowledge from \textit{in-domain} corpus, our proposed \textsc{RACo} leverages a much larger and general commonsense corpus collected from multiple sources that provide supportive evidences for various commonsense reasoning tasks. Meanwhile, we proposed several novel strategies for training a commonsense retriever that can be generalized to different commonsense reasoning tasks.

\section{Proposed Method}
\label{sec:method}
\begin{figure}
  \begin{center}
    \includegraphics[width=0.485\textwidth]{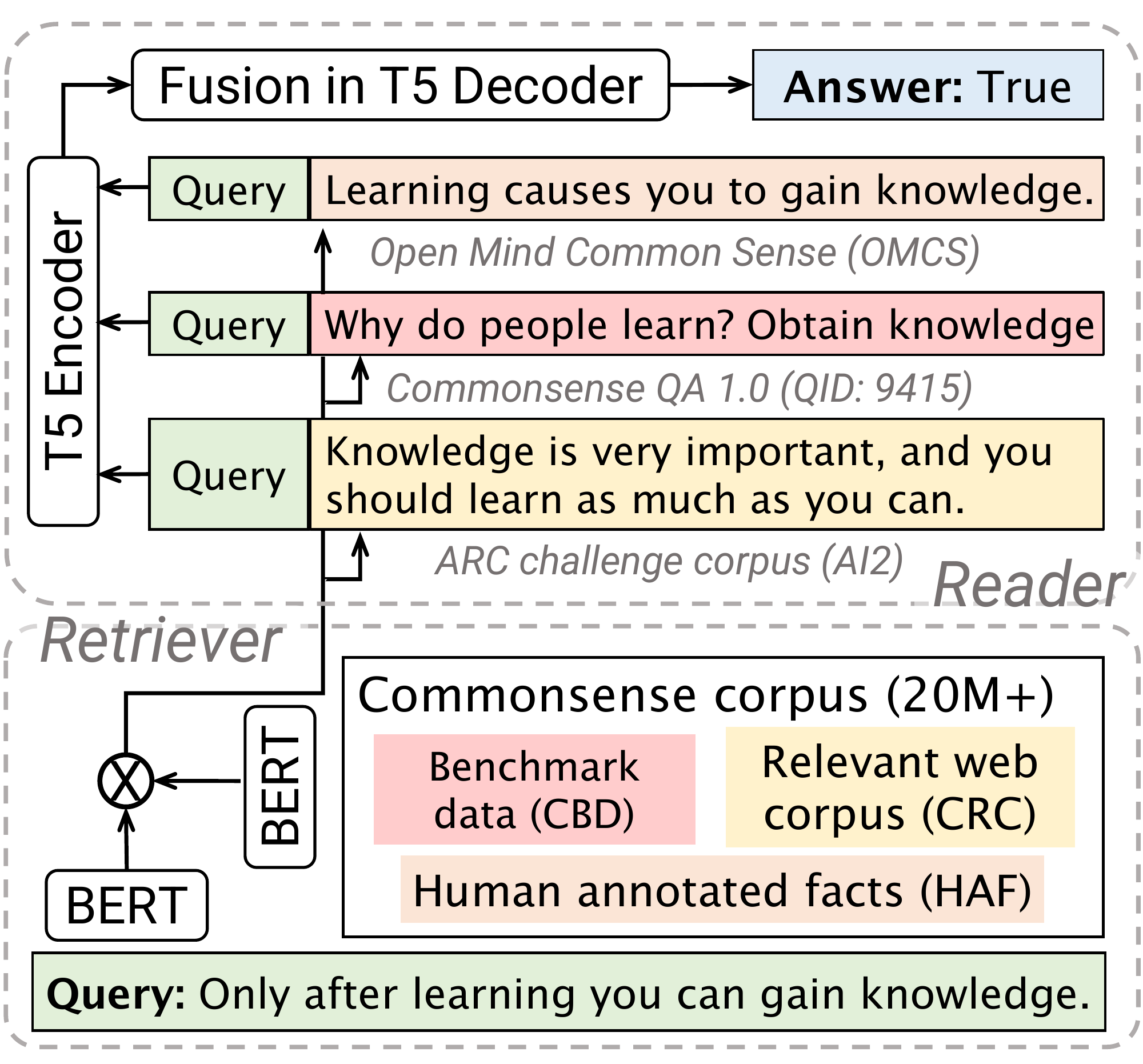}
  \end{center}
  \vspace{-0.15in}
  \caption{\textsc{RACo} has two major components: (i) a document retriever and (ii) a document reader. Specifically, the document retriever aims to fetch a handful of relevant documents from a large document collections. The document reader takes the input text, as well as the support documents to produce the desired output.}
  \label{fig:Framework}
\end{figure}

In this section, we elaborate on how to leverage commonsense knowledge retrieval from a large-scale corpus to improve various commonsense reasoning tasks, including commonsense corpus construction ($\S$\ref{sec:cscorpus}), commonsense document retriever ($\S$\ref{sec:retrieval}) and commonsense document reader ($\S$\ref{sec:reader}). The architecture of \textsc{RACo} is shown in Figure \ref{fig:Framework}.




\subsection{Commonsense Corpus Construction}
\label{sec:cscorpus}

Commonsense knowledge includes the basic facts about situations in everyday life, which is shared by most people and implicitly assumed in communications~\cite{li2022survey}.
Commonsense knowledge has two important properties: \textit{large} and \textit{diverse}.

Regarding the scale of knowledge, many commonsense corpus contains millions of statements. For example, Wiktionary
has more than one million word definitions and descriptions in English. 
Meanwhile, the commonsense knowledge is diverse, involving various aspects including human interaction and object properties. 
For example, OMCS\footnote{ \url{https://en.wikipedia.org/wiki/OMCS}} covers multiple domains of commonsense such as everyday events and
their effects (e.g., mop up the floor if we split food over it), facts about beliefs and desires (e.g., study hard to win scholarship), and properties of objects (e.g., goat has four legs). 
The diversity of knowledge is beneficial for retrieval-augmented methods because it enables relevance comparison across different sources, and offers textual knowledge to easily augment the input of generation models by concatenation. 
To build a large-scale commonsense corpus covering diverse sources, we collected commonsense documents from the following three aspects: (i) human annotated facts; (ii) commonsense benchmark datasets; (iii) commonsense relevant web corpus. The statistics can be found in Table \ref{tab:stats}.

\begin{table}[t]
\begin{center}
\setlength{\tabcolsep}{3mm}{\scalebox{0.93}{\begin{tabular}{l|r|c}
\toprule
Corpus & \# Instance & Avg. Word \\
\midrule
HAF-corpus & 3,561,762 & 11.06 $\pm$ 5.86 \\
CBD-corpus & 2,881,609 & 12.78 $\pm$ 9.31 \\
CRC-corpus & 14,587,486 & 17.76 $\pm$ 10.4\\
\bottomrule
\end{tabular}}}
\end{center}
\vspace{-0.15in}
\caption{Statistics for the commonsense corpus. The total size of these corpora exceeds 20M documents.}
\label{tab:stats}
\end{table}

\vspace{0.03in}
\noindent\textbf{Human annotated facts (HAF).} It contains factual commonsense either annotated by human annotators or written by domain experts, including OMCS~\cite{havasi2010open}, ATOMIC~\cite{sap2019atomic}, Wiktionary~\cite{meyer2012wiktionary}.

\vspace{0.03in}
\noindent\textbf{Commonsense benchmark datasets (CBD).} It includes training data from 19 commonsense benchmark datasets, such as $\alpha$-NLI~\cite{bhagavatula2020abductive}. See Appendix \ref{sec:corpus} for more details.

\vspace{0.03in}
\noindent\textbf{Commonsense relevant corpus (CRC).} It consists of raw statements about commonsense from the web, usually after some simple filtering. We obtained a filtered version from AI2 commonsense corpus, which is a merged corpus collected from ARC~\cite{clark2018think}, QASC~\cite{khot2020qasc} and GenericsKB~\cite{bhakthavatsalam2020genericskb}.

\subsection{Commonsense Document Retrieval}
\label{sec:retrieval}

Given a collection of $M$ commonsense documents, the goal of our retriever is to map all the documents in a low-dimensional vector, such that it can efficiently retrieve the top-$k$ documents relevant to the input text. Note that $M$ can be very large (e.g., over 20 million in our experiments) and $k$ is usually small (e.g., 10 or 20 in our experiments).

In this work, we follow the neural document retriever DPR~\cite{karpukhin2020dense} to employ two independent BERT~\cite{devlin2019bert} models to encode the query and the document separately, and estimate their relevance by computing a single similarity score between their $\text{[CLS]}$ token representations. 
Specifically, the document encoder $E_D(\cdot)$ which maps any text document to a low-dimensional real-valued vectors and builds an index for all the $M$ documents used for retrieval. At runtime, it applies a different query encoder $E_Q(\cdot)$ that maps the input question to a vector of the same dimension as the document vector, and retrieves top-$k$ documents of which vectors are the closest to the question vector. The similarity between the question and the document is calculated by the dot product of their vectors.

\vspace{-0.1in}
\begin{equation}
    \mathrm{sim}(q, d) = E_Q(q)^\mathsf{T} E_D(d).
\end{equation}



Recent efforts have shown that DPR transfer poorly to other domains~\cite{li2021encoder,kulshreshtha2021back}.
Thus, the primary challenge of training a strong commonsense retriever is to appropriately construct positive pairs and hard negative pairs~\cite{karpukhin2020dense,xiong2021approximate}. 
To do this, we propose novel strategies to construct question-document pairs that can be used for training a strong commonsense retriever.

\subsubsection{Positive Training Pairs}

In open-domain document retrieval, it is often the case that positive training pairs are available explicitly. For example, DPR treated Wikipedia documents that contain the correct answer as positive documents~\cite{karpukhin2020dense}. However, such training pairs might not be applicable on commonsense reasoning tasks because the output (e.g., \textit{True / False} in a binary question answering) is not supposed to be a sub-string of retrieved documents.

In order to train a strong commonsense dense retriever, we propose two novel strategies to construct positive training pairs, as described below.

\vspace{0.05in}
\noindent\textbf{Explanation as positive document.} 
The first method for constructing positive training pairs is to take human annotated explanations as positive documents. 
For examples, taking the question \textit{``Where do people go to pray? (A) church''} from CommonsenseQA1.0 as input, the explanation annotated in~\citet{aggarwal2021explanations} is \textit{``People go to a church to pray''}; similarly, a positive document for the question \textit{``When food is reduced in the stomach, nutrients are being deconstructed''} in OpenBookQA~\cite{mihaylov2018can} could be \textit{``Digestion is when stomach acid breaks down food''}.
The explanations have two important properties. First, they contain commonsense knowledge, such as \textit{people praying in church}, in the form of natural language. Second, they can be used to help select the correct choice in commonsense reasoning tasks.
So, we take advantage of the high correlation of natural language explanations with the input query, defining the input query as $q$ and the corresponding generated explanation as $d$ to train the retriever.

\vspace{0.05in}
\noindent\textbf{Ground truth output as positive document.} The second method for constructing positive training pairs is to directly use ground truth outputs in generation tasks as positive documents.
The ground truth output can be seen as a natural positive document that the retriever should retrieve. 
For example, in the CommonGen~\cite{lin2020commongen} dataset, the ground truth output for an input concept set \textit{\{dance, kid, room\}} is \textit{``a group of kids are dancing around a living room''}.
We define the input sequence in a generation task as $q$ and its corresponding ground truth output as $d$ to train the retriever.
During training, the vector distance between them are minimized.
During inference, though the ground truth documents are no longer in the commonsense corpus, the retriever can still fetch relevant documents similar to the ground truth output such as \textit{``a couple of kids are dancing on the floor (ARC corpus)''}, which provides relevant contexts describing the potential reaction between the input concepts \textit{``kid''} and \textit{``dance''}, hence helps generate desired outputs.

\subsubsection{Negative Training Pairs}

For negative pairs, we adopt the trick of in-batch negatives, which has been shown as an effective strategy for learning a dual-encoder model and used in the many recent dense retrieval models~\cite{lee2019latent,karpukhin2020dense}.

\subsection{Commonsense Document Reader}
\label{sec:reader}

After retrieving commonsense documents, the reader takes the input text along with the retrieved documents to produce the desired output. Sequence classification tasks are considered as a target sequence of length one. In our work, we use the fusion-in-decoder (FiD)~\cite{izacard2021leveraging} model as the reader.
Specifically, each retrieved document is concatenated with the input text, then independently encoded by the T5~\cite{raffel2020exploring} encoder.
Then, the T5 decoder performs cross-attention over the concatenation of the resulting representations of all the retrieved documents.

\section{Experiments}
\label{sec:Experiments}

\begin{table*}[t]
\begin{center}
\setlength{\tabcolsep}{2.2mm}{\scalebox{0.88}{\begin{tabular}{l||c|ccccc|ccccc}
\toprule
{\multirow{2}*{Methods $\downarrow$}} & {\multirow{2}*{K-type}} & \multicolumn{5}{c|}{{CommonGen}} & \multicolumn{5}{c}{{ComVE}} \\
&  & BL-4 & RG-L & MET & CIDEr & SPICE$^{*}$ & BL-4$^{*}$ & RG-L & MET & CIDEr & SPICE \\ 
\midrule
BART-large & - & 26.30 & 41.98 & 30.90 & 13.92 & 30.60 & 19.22 & 44.86 & 27.10 & 11.04 & 35.14 \\ 
T5-large & - & 28.60 & 42.97 & 30.10 & 14.96 & 31.60 & 22.77 & 51.83 & 26.66 & 11.42 & 34.62 \\
CALM & CLM & 29.50 & - & 31.90 & 15.61 & 33.20 & 23.50 & 52.56 & 27.41 & 11.87 & 35.23 \\ 
\textsc{Unicorn} & CLM & 39.86 & 44.56 & 34.52 & 17.26 & 30.20 & 24.46 & 52.75 & 27.88 & 12.40 & 35.79 \\ 
KG-BART & KGR & 30.90 & 44.54 & 32.40 & 16.83 & 32.70 & - & - & - & - & - \\
GraphRF & KGR & - & - & - & - & - & 22.07 & 44.32 & 25.96 & 11.62 & 33.09 \\
MoKGE & KGR & - & - & - & - & - & 22.87 & 52.03 & 27.01 & 11.75 & 34.88 \\ 
KFCNet & RAM & 41.97 & 46.13 & 36.22 & 17.39 & 33.11 & - & - & - & - & - \\ 
BM25+FiD & RAM & 42.17 & 47.95 & 35.57 & 18.74 & 33.16 & 24.39 & 52.76 & 28.26 & 12.67 & 35.28 \\ 
\textsc{RACo} & RAM & \textbf{42.76} & \textbf{48.19} & \textbf{35.80} & \textbf{18.89} & \textbf{33.89} & \textbf{25.30} & \textbf{53.29} & \textbf{28.62} & \textbf{12.76} & \textbf{36.37} \\ 
\bottomrule
\end{tabular}}}
\end{center}
\vspace{-0.18in}
\caption{Compared with commonsense-aware language models (CLM) and knowledge graph reasoning models (KGR) counterparts, our retrieval-augmented commonsense reasoning (\textsc{RACo}) can outperform the baseline methods and achieved state-of-the-art performance on the CommonGen and ComVE benchmarks. *: primary metric.}
\label{tab:baseline-nlg}
\end{table*}

\begin{table}[t]
\begin{center}
\setlength{\tabcolsep}{3.1mm}{\scalebox{0.88}{\begin{tabular}{l||c|c|c}
\toprule
{\multirow{2}*{Methods $\downarrow$}} & {\multirow{2}*{K-type}} & CSQA1.0 & OBQA \\
& & ACC. & ACC. \\ 
\midrule
T5-large & - & 70.14 & 66.02 \\
\textsc{Unicorn} & CLM & 71.60 & 70.02 \\
KagNet & KGR & 69.00 & -  \\ 
QA-GNN & KGR & 73.40 & 67.80 \\
GreaseLM & KGR & 74.20 & 66.90 \\
BM25+FiD & RAM & 74.12 & 67.75 \\
\textsc{RACo} & RAM & \textbf{75.76} & \textbf{71.25} \\
\bottomrule
\end{tabular}}}
\label{tab:baseline-mcqa}
\end{center}
\vspace{-0.18in}
\caption{RACo achieves better performance than other knowledge-enhanced method counterparts.}
\end{table}

\begin{table}[t]
\begin{center}
\setlength{\tabcolsep}{3.1mm}{\scalebox{0.88}{\begin{tabular}{l||c|c|c}
\toprule
{\multirow{2}*{Methods $\downarrow$}} & {\multirow{2}*{K-type}} & CSQA2.0 & CREAK \\
& & ACC. & ACC. \\ 
\midrule
T5-large & - & 54.60 & 77.32 \\
\textsc{Unicorn} & CLM & 54.90 & 79.51 \\
GreaseLM & KGR & - & 77.51 \\
BM25+FiD & RAM & 58.75 & 83.03 \\
\textsc{RACo} & RAM & \textbf{61.75} & \textbf{84.17} \\
\bottomrule
\end{tabular}}}
\end{center}
\vspace{-0.18in}
\caption{\textsc{RACo} outperforms the baseline methods and achieved state-of-the-art performance on the CREAK.}
\label{tab:baseline-ccv}
\end{table}

\begin{table}[t]
\begin{center}
\setlength{\tabcolsep}{2mm}{\scalebox{0.9}{\begin{tabular}{l||cc||cc}
\toprule
{\multirow{2}*{\makecell[l]{Retriever \\ in \textsc{RACo} $\downarrow$ }}} & \multicolumn{2}{c|}{{CSQA1.0}} & \multicolumn{2}{c}{{OBQA}} \\
& Hit@5 & Hit@10 & Hit@5 & Hit@10 \\
\midrule
BM25 & 46.33 & 50.93 & 50.21 & 60.24 \\
DPR$_{\text{Wiki}}$ & 4.40 & 5.78 & 28.41 & 37.58 \\
DPR$_{\textsc{RACo}}$ & \textbf{61.71} & \textbf{65.68} & \textbf{75.23} & \textbf{85.41} \\  
\bottomrule
\end{tabular}}}
\end{center}
\vspace{-0.18in}
\caption{Retrieval accuracy on dev sets, measured as the percentage of retrieved documents that contain the ground truth document, which is annotated in~\citet{mihaylov2018can,aggarwal2021explanations}.
In addition, DPR$_{\text{Wiki}}$ directly uses the DPR trained on Wikipedia for commonsense retrieval without any fine-tuning process. 
DPR$_{\text{RACo}}$ trains the commonsense dense retrieval using our proposed training pairs construction strategy.
}
\label{tab:hit}
\end{table}

\subsection{Tasks and Datasets}
\label{sec:tasks}

\vspace{0.05in}
\noindent\textbf{Multi-choice question answering.}
Give a question, an intelligent system is asked to select one correct answer from the choices offered as a list. 
We conducted experiments on  CommonsenseQA1.0~\cite{talmor2019commonsenseqa} and OpenBookQA~\cite{clark2018think}.
CommonsenseQA1.0 (CSQA1.0) contains 12,102 questions with one correct answer and four distractor answers. OpenBookQA (OBQA) consists of 5,957 elementary-level questions with one correct answer and three distractor answers.
For evaluation, the primary metric on these two tasks is accuracy (ACC.).

\vspace{0.05in}
\noindent\textbf{Commonsense fact verification.}
Given a commonsense claim, an intelligent system is expected to verify the statement in natural text against facts. For example, the statement \textit{"A pound of cotton has the same weight as a pound of steel"} in the CommonsenseQA2.0~\cite{talmor2021commonsenseqa} should be identified as \textit{true}. We conducted experiments on two commonsense fact verification datasets, including CommonsenseQA2.0~\cite{talmor2021commonsenseqa} and CREAK~\cite{onoe2021creak}.
CommonsenseQA2.0 was collected via gamification, which includes 14,343 assertions about everyday commonsense knowledge. CREAK is designed for commonsense reasoning about entity knowledge, which consists of 13,000 assertions about entities. For evaluation, the primary metric is accuracy (ACC.).

\vspace{0.05in}
\noindent\textbf{Constrained commonsense generation.}
Given a set of concepts such as 
\textit{``dog, frisbee, catch, throw''}, the task is to generate a coherent sentence describing an everyday scenario such as \textit{``a man
throws a frisbee and his dog catches it''}.
Our experiments were conducted on the benchmark dataset Commongen~\cite{lin2020commongen}. It consists of 79,000 commonsense descriptions over 35,000 unique concept-sets. The average input / output length is 3.4 / 10.5 words. All examples in the dataset have 4-6 references. The task is evaluated by SPICE~\cite{anderson2016spice}, BLEU-4~\cite{papineni2002bleu}, ROUGE-L~\cite{lin2004rouge}, CIDEr~\cite{vedantam2015cider}, in which SPICE is the primary metric for leaderboard ranking.

\vspace{0.05in}
\noindent\textbf{Counterfactual explanation generation.}
Given a counterfactual statement, the task is to generate reasons why the statement does not make sense. Our experiments were conducted on the benchmark dataset ComVE from SemEval-2020 Task 4~\cite{wang2020semeval}. It contains 11,997 examples. The average input/output length is 7.7 / 9.0 words. All ground truth have 3 references. The task is evaluated by SPICE~\cite{anderson2016spice}, BLEU-4~\cite{papineni2002bleu}, ROUGE-L~\cite{lin2004rouge}, CIDEr~\cite{vedantam2015cider}, in which BLEU-4 is the primary metric for leaderboard ranking.

\subsection{Baseline Methods}
\label{sec:baseline}

We compared our \textsc{\textsc{RACo}} with various kinds of baseline methods. In addition of comparing with pre-trained language models, such as BART~\cite{lewis2020bart} and T5~\cite{raffel2020exploring}, we also compared with three kinds of commonsense knowledge augmented methods as introduced below.  

\vspace{0.03in}
\noindent\textbf{Commonsense-aware language models (CLM).} 
They are trained with external commonsense corpus or datasets, in order to embed commonsense knowledge into their parameters.
During fine-tuning, the language models make predictions without accessing to any external corpus. 
In the experiments, we compared our model with CALM~\cite{zhou2021pre} and \textsc{Unicorn}~\cite{lourie2021unicorn}.

\vspace{0.03in}
\noindent\textbf{Knowledge graph reasoning models (KGM).} KGs are incorporated into models for augmenting the limited information in the input texts. 
We compared our model with KagNet~\cite{lin2019kagnet}, GRF~\cite{ji2020language}, KG-BART~\cite{liu2021kg}, QA-GNN~\cite{yasunaga2021qa}, MoKGE~\cite{yu2022diversifying} and GreaseLM~\cite{zhang2022greaselm}.

\vspace{0.03in}
\noindent\textbf{Retrieval augmented models (RAM).} We compared with a recent retrieval-augmented method named KFCNet~\cite{li2021kfcnet} for constraint commonsense generation. 
In addition, we also compared with using sparse retriever such as BM25 to retrieve knowledge from our constructed commonsense corpus and use FiD~\cite{izacard2021leveraging} as generator to produce outputs.

\begin{table*}[t]
\begin{center}
\setlength{\tabcolsep}{3    mm}{\scalebox{0.87}{\begin{tabular}{l||c|c|c|c|cc|cc|c}
\toprule
Retriever & CSQA1.0 & OBQA & CSQA2.0 & CREAK & \multicolumn{2}{c|}{{CommonGen}} & \multicolumn{2}{c|}{{ComVE}} & {\multirow{2}*{Avg.}} \\
training set $\downarrow$ & ACC. & ACC. & ACC. & ACC. & BL-4 & SPICE$^{*}$ & BL-4$^{*}$ & SPICE & \\
\midrule
OBQA & 71.09 & 69.55 & 57.57 & 81.47 & 33.70 & 33.83 & 27.55 & 38.29 & 58.67 \\
CommonGen & 73.36 & 66.34 & 57.49 & 83.15 & \textbf{38.34} & \textbf{36.71} & 27.71 & 37.79 & 59.46 \\
CSQA1.0 & 74.84 & \textbf{71.56} & 59.76 & 82.85 & 35.29 & 35.05 & 27.28 & 35.05 & 60.50 \\
All datasets & \textbf{75.08} & 70.40 & \textbf{60.21} & \textbf{84.01} & 37.26 & 36.03 & \textbf{28.05} & \textbf{38.17} & \textbf{60.80} \\
\bottomrule
\end{tabular}}}
\end{center}
\vspace{-0.18in}
\caption{Model performance (on dev sets) of using commonsense retrievals trained on different datasets. Training with question-document pairs from all datasets yield the best average performance on six tasks. *: primary metric.}
\label{tab:retriever}
\end{table*}

\begin{table}[t]
\begin{center}
\setlength{\tabcolsep}{2mm}{\scalebox{0.88}{\begin{tabular}{l|ccc|cc}
\toprule
Corpus & \multicolumn{3}{c|}{{Corpus name}} & \multicolumn{2}{c}{{Dataset name}} \\
size $\downarrow$ & HAF & CBD & CRC & CSQA2.0 & CREAK \\
\midrule
\multicolumn{1}{c|}{{0}} & \multicolumn{3}{c|}{{(no retrieval)}} & 53.80 & 77.32 \\
3.56M & $\checkmark$ & & & 56.57 & 78.91 \\
2.88M & & $\checkmark$ & & 56.65 & 78.93 \\
13.6M & & & $\checkmark$ & 56.86 & 82.82 \\
6.44M & $\checkmark$ & $\checkmark$ & & 57.46 & 79.09 \\
17.1M & $\checkmark$ & & $\checkmark$ & 58.16 & 82.52 \\
16.5M & & $\checkmark$ & $\checkmark$ & 58.39 & 82.98 \\
20.1M & $\checkmark$ & $\checkmark$ & $\checkmark$ & \textbf{59.66} &  \textbf{83.85} \\
\bottomrule
\end{tabular}}}
\end{center}
\vspace{-0.15in}
\caption{Performance on dev sets of retrieving commonsense knowledge from different size of corpus.}
\label{tab:corpus-size}
\end{table}

\subsection{Automatic Evaluation}

\subsubsection{\textsc{RACo} v.s. Baseline Methods}

\vspace{0.03in}
\noindent\textbf{Comparison with non-retrieval methods.} To observe the effectiveness of retrieval on commonsense reasoning tasks, we first compared model performance with and without commonsense retrieval. As shown in Table \ref{tab:baseline-nlg}-\ref{tab:baseline-ccv}, compared with BART and T5 that directly encode the input query and produce output without using external knowledge, our proposed \textsc{RACo} can improve the commonsense reasoning performance by a large margin. Specifically, \textsc{RACo} improved BLEU-4 by +8.44\% on the commonsense generation tasks, improved accuracy by +5.43\% on the multiple choice question answering tasks, and improved accuracy by +6.15\% on the commonsense verification tasks. Therefore, we concluded that \textsc{RACo} can leverage the retrieval of relevant references from commonsense corpora to help language models produce better outputs in various commonsense reasoning tasks.

\vspace{0.03in}
\noindent\textbf{Comparison with other knowledge-enhanced methods.} As mentioned in the $\S$\ref{sec:baseline}, the commonsense reasoning ability of a language model can be enhanced by fine-tuning on commonsense corpora or reasoning over multi-hop relations on knowledge graphs. 
As indicated by Table \ref{tab:baseline-nlg}-\ref{tab:baseline-ccv}, compared with commonsense-aware language models (CLM), retrieval augmented model explicitly leverage relevant commonsense knowledge, demonstrating superior performance on all datasets. Compared with knowledge graph reasoning methods (KGR), it can achieve better performance on all six datasets.

\subsubsection{Effects on Commonsense Retriever}
\label{sec:retriever}

To evaluate the effectiveness of commonsense retrieval, we compare the performance of different retriever training settings, including BM25, DPR$_{\text{Wiki}}$, and DPR$_{\text{RACo}}$. 
Specifically, DPR$_{\text{Wiki}}$ directly uses the DPR trained on Wikipedia for commonsense retrieval without any fine-tuning process. 
DPR$_{\text{RACo}}$ trains the commonsense dense retrieval using our our proposed training pairs construction strategy.
As shown in Table \ref{tab:hit}, we can observe DPR$_{\text{Wiki}}$ performs the worst among all retrievers. Our proposed DPR$_{\text{RACo}}$ can significantly improve the retrieval performance, compared to BM25. 
It is important to note that the performance of retrieval is not necessarily linearly related to the performance of final output. However, in general, the more relevant the retrieved content, the more helpful it is to produce better outputs during the reading step. The observation can also be drawn from the comparison of BM25+FiD and \textsc{RACo} in Tables \ref{tab:baseline-nlg}-\ref{tab:baseline-ccv}.

\subsubsection{Effects on Multi-dataset Training}

As shown in Table \ref{tab:retriever}, we compare the model performance of retrievers trained by different set of question-document pairs.
For example, the first line represents the retriever is trained with only question-document pairs (5,000 in total) from the OBQA dataset. The last line represents using question-document pairs from all six datasets.
From the table, we can observe when the retriever is trained on only one dataset, it might not work well on other datasets because of differences in data distribution.
Instead, training with multiple datasets demonstrates better generalizability. 

\subsubsection{Effects on Commonsense Corpus}

To validate the effect of the number and content of corpora on our proposed method, we test the corresponding model performance under different corpora, including choosing a corpus, or any combination of multiple corpora. In Table \ref{tab:corpus-size}, we show the performance of CSQA2.0 and CREAK on different commonsense corpora.
It is worth noting that compared with other data, CSQA2.0 and CREAK can more realistically reflect the impact of different corpora on model performance, 
mainly because these two datasets are \textit{not} based on any commonsense knowledge source during the collection process, so the coverage of the problem is much wider than other four datasets that are collects from a certain knowledge source. For example, CSQA1.0 and CommonGen are collected based on ConceptNet.

\subsubsection{Effects on Number of Documents}

\begin{figure}[t]
	\centering
	{\includegraphics[width=0.49\textwidth]{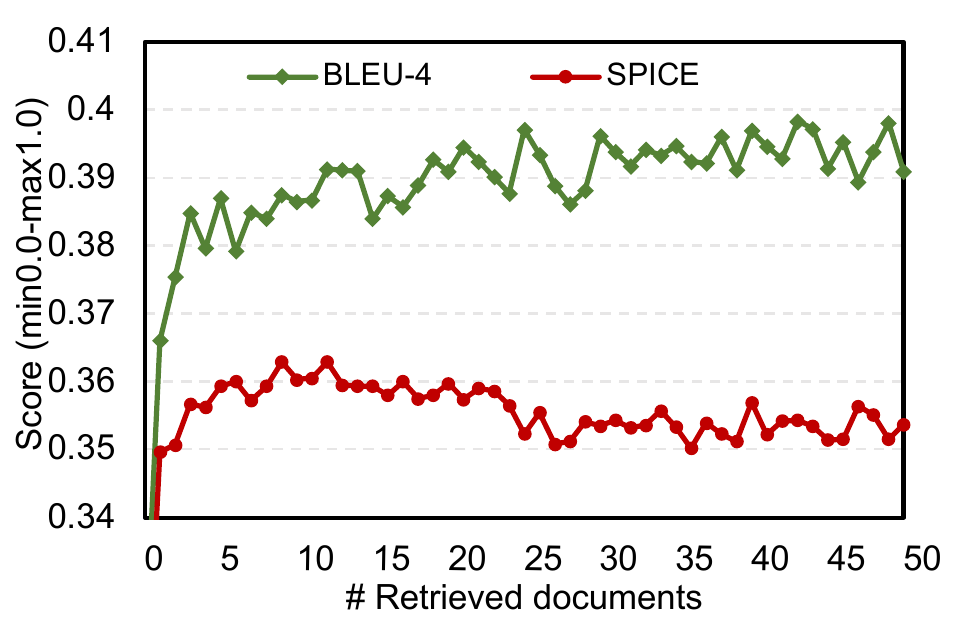}}
	\vspace{-0.25in}
	\caption{As the number of retrieved documents increases, the model performance of \textsc{RACo} on the CommonGen dataset first increases and then remains unchanged on BLEU-4 or even decreases on SPICE (the primary metric on the CommonGen leaderboard).}
	\label{fig:ktvsft}
\end{figure}

We also compared model performance with different numbers of retrieved documents. As shown in Figure \ref{fig:ktvsft}, as the number of retrieved documents increases, the model performance of RACO on the CommonGen dataset first increases and then remains unchanged on BLEU-4 or even decreases on SPICE (the primary metric on the CommonGen leaderboard), but the GPU memory consumption increases significantly. 
This is mainly because when the number of retrieved documents increases, more noisy information might be included in the model, which could hurt the performance of the model. Thus, with reasonable computational overhead, we only use 10 documents in our experiments.

\begin{table}[t]
\begin{center}
\setlength{\tabcolsep}{1.8  mm}{\scalebox{0.88}{\begin{tabular}{l|c|c|c|c}
\toprule
& \multicolumn{2}{c|}{{CommonGen}} & \multicolumn{2}{c}{{ComVE}} \\
& Fluency & Accuracy & Fluency & Accuracy \\ 
\midrule
T5 & 2.86 & 2.78 & 2.94 & 2.74 \\ 
CALM & 2.85 & 2.81 & 2.93 & 2.78 \\
\textsc{Unicorn} & 2.88 & 2.84 & \textbf{2.95} & 2.78 \\ 
\textsc{RACo} & \textbf{2.90} & \ \ \textbf{2.96}* & 2.94 & \ \ \textbf{2.86}* \\
\bottomrule
\end{tabular}}}
\end{center}
\vspace{-0.15in}
\caption{Human evaluations by independent scoring based on accuracy and fluency. * indicates p-value < 0.05 under paired t-test between \textsc{RACo} and baselines.}
\label{tab:human}
\end{table}

\subsection{Human Evaluation}

We randomly sample 50 generated outputs from the CommonGen dev set (as the test set is not public) and 50 generated outputs from the ComVE test set. All evaluations were conducted on Amazon Mechanical Turk (AMT), and each evaluation form was answered by three AMT workers.
The generated outputs are evaluated by \textit{fluency} and \textit{accuracy}. 
Fluency is assessed on the grammatical correctness and readability of the generated outputs disregarding the input text. 
Besides, accuracy evaluates whether the output generated is correct and reasonable given the input text of each task.

As shown in Table \ref{tab:human}, our model significantly outperforms baseline methods in terms of accuracy and fluency on both datasets.
In particular, the accuracy of the generated output is greatly improved due to the incorporation of the retrieved commonsense knowledge.
Furthermore, since all baseline models are pre-trained on large-scale corpora, they all produce outputs with great fluency. 
However, compared with baseline methods, the outputs generated by our model on the CommonGen dataset still have better fluency. 
This is mainly because the retrieved references are semantically complete sentences with good fluency, which might mitigate grammatical errors during the generation process.

\begin{table}[t]
\centering
\scalebox{0.88}{\begin{tabular}{p{8.3cm}}
\toprule
\textbf{1. CSQA2.0 Statement --} A private college is usually smaller than a public university in attendance. (\textbf{True}) \\
\textbf{Retrieved evidence -- \#1} Private schools are usually small and are worth the cost. \textbf{\#2} University's are larger then most colleges. \textbf{\#3} Colleges considered ``small'' have fewer than 5,000 students \\
\midrule
\textbf{Predictions --} \textbf{T5 and \textsc{Unicorn}:} False \textbf{\textsc{RACo}:} True \\
\midrule
\midrule
\textbf{2. ComVE Statement --} The sun made my t-shirt wet. \\
\textbf{Retrieved evidence -- \#1} The sun can dry wet clothes. \textbf{\#2} The sun can dry something that is wet. \\
\midrule
\textbf{Generated outputs --} \textbf{T5:} The sun is hot. \textbf{\textsc{Unicorn}:} The sun does not make clothes wet. \textbf{\textsc{RACo}:} The sun would dry a t-shirt but not make a t-shirt wet. \\
\midrule
\midrule
\textbf{3. Commongen Input Words --} eye look move \\
\textbf{Retrieved evidence -- \#1} She moves her eyes around. \textbf{\#2} The eye looks towards the peaks. \textbf{\#3} A woman looks at the camera as she moves each eye individually. \textbf{\#4} His eyes move across the paper. \\
\midrule
\textbf{Generated outputs --} \textbf{T5:} Someone looks at someone, then moves his eyes. \textbf{\textsc{Unicorn}:} Someone looks at her and moves her eyes.	\textbf{\textsc{RACo}:} A man moves his eyes to look at the camera.  \\
\bottomrule
\end{tabular}}
\vspace{-0.1in}
\caption{Case study. 
RACo corrects the erroneous predictions of baseline models (e.g., T5 and \textsc{Unicorn}) using the retrieved commonsense knowledge.
}
\label{tab:case-study}
\end{table}

\subsection{Case Study}

Table \ref{tab:case-study} shows two examples with predictions from different models. 
We demonstrate a ``comparison'' statement from CSQA2.0 as the first example. As shown in the table, both T5 and \textsc{Unicorn} make wrong predictions, demonstrating a lack of commonsense knowledge. By leveraging the retrieved evidence from commonsense corpus, our proposed \textsc{RACo} can tell the statement ``private college is usually smaller than a public university in attendance'' is true.
In addition, we show an example from counterfactual explanation generation task as the second example. Among the three outputs shown, the explanation generated by T5 is less associated with the input statement. Compared with the generated outputs from \textsc{Unicorn}, our model can generate a semantically richer and more reasonable explanation. This is mainly because the references retrieved provide strong evidence from the perspective of the sun dries things out.


\section{Epilogue}
\label{sec:conclusions}

\vspace{0.05in}
\noindent\textbf{Conclusions.} Retrieval-augmented methods have been widely used in many NLP tasks such as open-domain question answering. 
However, applying this approach to commonsense reasoning has been neglected in the existing literature. 
In this paper, we systematically investigate how to leverage commonsense knowledge retrieval to improve commonsense reasoning tasks.
We collected a commonsense corpus containing over 20 million documents, and proposed novel strategies for training a commonsense retriever. 
Extensive experiments demonstrate our method can effectively improve the performance of various commonsense reasoning tasks, achieving new state-of-the-art performance on the CommonGen and CREAK leaderboards.

\vspace{0.05in}
\noindent \textbf{Future work.} A natural extension of this work is to leverage heterogeneous knowledge to improve commonsense reasoning tasks, such as combining structured (i.e., knowledge graph) and unstructured (i.e., retrieved text) knowledge. Such a model will require building a graph reasoning module and a textual reasoning module, and merging the knowledge learned from both, which is a challenging task. 
The second future direction is to learn a commonsense dense retriever without question-document pairs. 
For example, in binary question answering, the labels are \textit{True} / \textit{False} that cannot be used to train a commonsense retriever.

\vspace{0.05in}
\noindent\textbf{Limitations.}
There are two main limitations.
First, \textsc{RACo} retrieves documents from a large-scale corpus, then leverage the retrieved documents to make predictions. So, compared with baseline methods such as T5 and \textsc{Unicorn}, \textsc{RACo} consumes more time and computing resources.
Second, due to the diversity and multi-source nature of commonsense knowledge, the retrieved evidence might contain noisy information that can even hurt model performance. 
A fine-grained filtering or re-ranking module could be a future work.

\section*{Acknowledgement}

This work was supported in part by NSF IIS-1849816, IIS-2119531, IIS-2137396, IIS-2142827, CCF-1901059, and ONR N00014-22-1-2507. 

\balance
\bibliography{reference}
\bibliographystyle{acl_natbib}

\clearpage
\balance
\appendix
\section{Appendix}
\subsection{Commonsense Retrieval Corpus}
\label{sec:corpus}

We use a combination of 19 commonsense datasets for our largest scale training data retrieval. The datasets include
$\alpha$-NLI~\cite{bhagavatula2020abductive}, SWAG~\cite{zellers2018swag}, RACE~\cite{lai2017race}, CODAH~\cite{chen2019codah}, CommonsenseQA1.0~\cite{talmor2019commonsenseqa}, 
CommonsenseQA2.0~\cite{talmor2021commonsenseqa}, WinoGrade~\cite{sakaguchi2021winogrande}, ARC~\cite{clark2018think}, CREAK~\cite{onoe2021creak}, OBQA~\cite{mihaylov2018can}, PhysicalIQA~\cite{bisk2020piqa}, QASC~\cite{khot2020qasc},
SocialIQA~\cite{sap2019social}, CosmosQA~\cite{huang2019cosmos},
MNLI~\cite{williams2018broad},
VATEX~\cite{wang2019vatex},
Activity~\cite{krishna2017dense},
SNLI~\cite{bowman2015large}
STSB~\cite{cer2017semeval}.

\subsection{Implementation Details}
\label{sec:imple-details}

\vspace{0.02in}
\noindent\textbf{Retriever.} We employed two independent pre-trained BERT-base models with 110M parameters~\cite{devlin2019bert} as query and document encoders.
BERT-base consists of 12 Transformer layers. For each layer, the hidden size is set to 768 and the number of attention head is set to 12. 
All dense retrievers were trained for 40 epochs with a learning rate of 1e-5. We used Adam~\citep{kingma2015adam} as the optimizer, and set its hyperparameter $\epsilon$ to $1e$-$8$ and $(\beta_1, \beta_2)$ to $(0.9, 0.999)$. The batch size is set as 32 on 8x32GB Tesla V100 GPUs.

\vspace{0.03in}
\noindent\textbf{Reader.} We employed the FiD~\cite{izacard2021leveraging} model that is built up on T5-large~\cite{raffel2020exploring}. 
For model training, we used AdamW~\cite{loshchilov2017decoupled} with batch size 32 on 8x32GB Tesla V100 or A100 GPUs.
We experimented with learning rates of 1e-5/3e-5/6e-5/1e-4 and we found that in general the model performed best when set to 3e-5. All reader models were trained with 20,000 steps in total where the learning rate was warmed up over the first 2,000 steps, and linear decay of learning rate.


\subsection{Additional Related Work}

Pre-training a language model on commonsense corpora is the most straightforward way to learn commonsense knowledge. Meanwhile, it also helps avoid overfitting when fine-tuned on downstream tasks.
When the commonsense materials are appropriately selected, this simple strategy could demonstrate significantly superior performance than vanilla pre-trained language models~\cite{zhou2021pre}.
Notable methods include COMET~\cite{bosselut2019comet}, CALM~\cite{zhou2021pre}, Unicorn~\cite{lourie2021unicorn} and etc. 
For example, COMET initialized its parameters from GPT-2 and post-trained on ATOMIC to adapt its learned representations to knowledge generation, and produces novel knowledge tuples that are high quality~\cite{bosselut2019comet}.
Unicorn initialized its parameters from T5 and performed a multi-task training on six commonsense question answering datasets~\cite{lourie2021unicorn}.
While this development is exhilarating, such commonsense-aware language models still suffer from the following drawbacks: 
first, they are usually trained offline, rendering the model agnostic to the latest information, e.g., Covid-19 is a disease caused by a coronavirus discovered in 2019.
Second, they make predictions by only ``looking up information” stored in its parameters, leading to inferior interpretability~\cite{shuster2021retrieval}.

Incorporating knowledge graph (KG) is essential for many commonsense reasoning tasks to augment the limited textual information. 
A KG often provides comprehensive and rich entity features and relations so models can easily traverse links to discover how entities are interconnected to express certain commonsense knowledge.
Some recent work explored using graph neural networks (GNN) to reason over multi-hop relational KG paths, yielding remarkable performance on many commonsense reasoning tasks, such as commonsense question answering~\cite{lin2019kagnet,yasunaga2021qa,zhang2022greaselm}, abductive reasoning~\cite{ji2020language,yu2022diversifying}, and chit-chat dialogue systems~\cite{zhou2018commonsense,zhang2020grounded}.
The most frequently used KG is ConceptNet. For example, \citet{ji2020language} enriched concept representations in the input text with neighbouring concepts on ConceptNet and performed dynamic multi-hop reasoning on multi-relational paths so the knowledge can be embedded into the hidden representations.
Nevertheless, the type of commonsense knowledge is restricted by the relations defined in a knowledge graph schema. Meanwhile, commonsense knowledge lies at an unbounded set of facts and situations that usually cannot be covered by a single knowledge graph. Reasoning over multiple knowledge graph is a challenging task.

\begin{figure}[t]
	\centering
	{\includegraphics[width=0.48\textwidth]{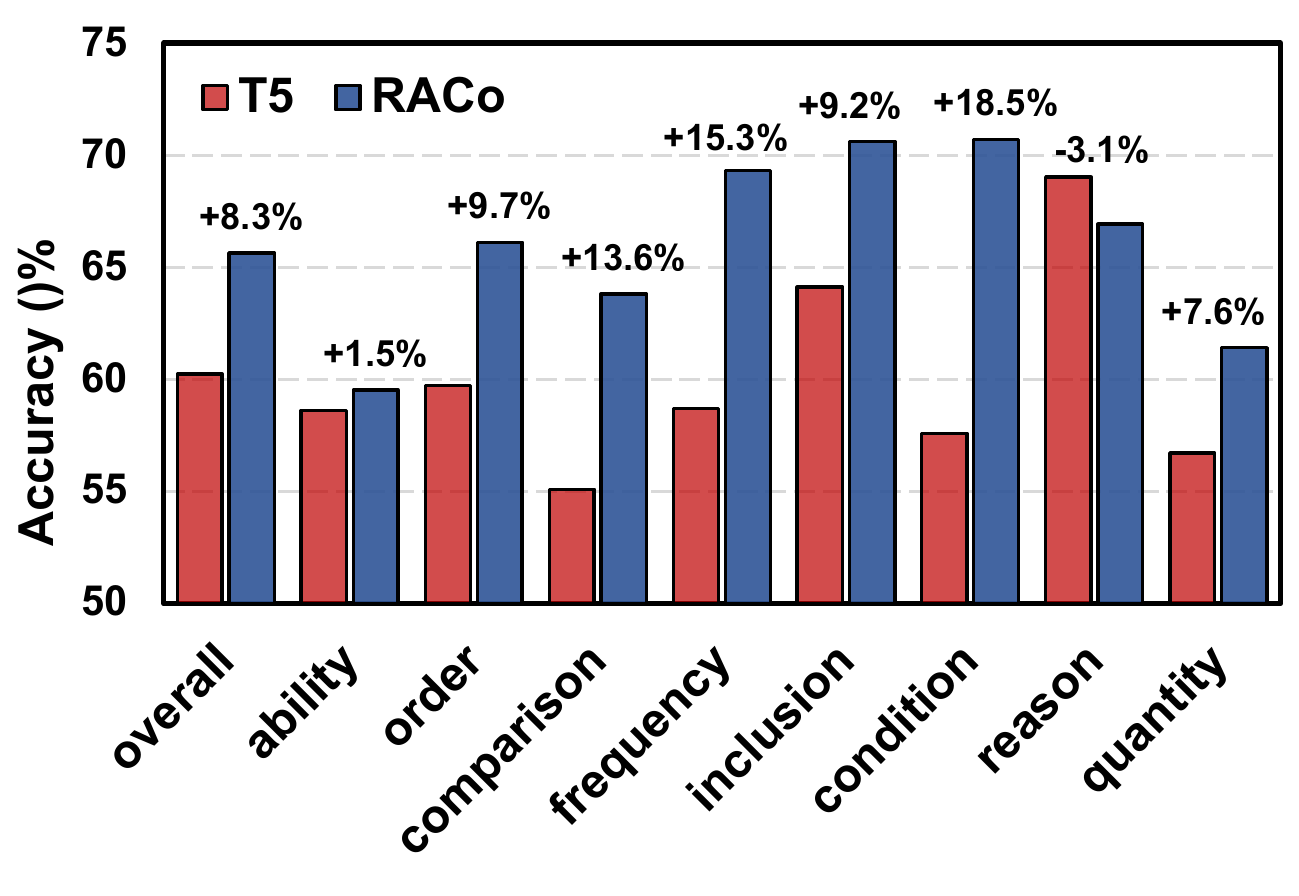}}
	\vspace{-0.3in}
	\caption{Performance of T5 and our \textsc{RACo} on different commonsense statement types in CSQA2.0.}
	\label{fig:case-csqa}
\end{figure}

\subsection{Case Study on CSQA2.0}

Figure \ref{fig:case-csqa} demonstrates the accuracy of T5 and our \textsc{RACo} for different statement types on the CSQA2.0 dataset. 
First, compared to T5, our model can improve by 8.3\% accuracy on all dev data (shown in the first column). However, on different statement types, the model performance is different. For example, from the predicted results of T5, the performance on "comparison" statements and "condition" statements is below-average.  
By introducing the retrieved commonsense knowledge, \textsc{RACo} demonstrated significantly better performance on these two sub-categories, achieving 15.3\% and 18.5\% improvement, which is significantly higher than the average 8.3\% improvement. Nevertheless, we also observe the retrieved evidence might provide noisy information, resulting in performance degradation, such as ``reason'' related statements. We show an example in Table \ref{tab:case-study}.
Statements under these categories are often descriptions or comparisons of factual commonsense, the retrieved documents can thus well complement the necessary knowledge of a given statement.
However, some statements require the model to reason in a given scenario, so making correct predictions requires the model to use commonsense knowledge to understand the local contexts. In these statements, retrieved knowledge might even contradict the assumptions, hurting the model performance.

\end{document}